\documentclass[10pt,twocolumn,letterpaper]{article}
\PassOptionsToPackage{table}{xcolor}
\usepackage[pagenumbers]{cvpr}

\makeatletter
\patchcmd{\paragraph}{3.25ex \@plus1ex \@minus.2ex}{2pt \@plus1pt \@minus1pt}{}%
  {\PackageError{manuscript}{Could not adjust paragraph heading spacing}{Check the paragraph definition.}}
\makeatother
\newcommand{\boldheading}[1]{\par\addvspace{2pt plus 1pt minus 1pt}\noindent\textbf{#1}}
\newcommand{\ablationheading}[1]{\par\addvspace{2pt}\noindent\textbf{#1}}
\makeatletter
\renewcommand{\cvprsection}{\@startsection{section}{1}{\z@}
  {-10pt}{7pt}{\large\bf}}
\renewcommand{\cvprsubsection}{\@startsection{subsection}{2}{\z@}
  {-8pt}{5pt}{\elvbf}}
\renewcommand{\cvprsubsubsection}{\@startsection{subsubsection}{3}{\z@}
  {-6pt}{3pt}{\tenbf}}
\makeatother

\usepackage{amssymb}
\usepackage{xspace}
\DeclareMathOperator{\clip}{clip}

\newcommand{\method}{AdaSeg\xspace}
\newcommand{\methodsource}[1]{\,{\scriptsize\textcolor{gray}{\textit{#1}}}}

\usepackage{multirow}
\usepackage{siunitx}

\usepackage{colortbl}
\usepackage{placeins}

\definecolor{captionGainPositive}{HTML}{3B6E9E}
\definecolor{captionGainNegative}{HTML}{A95C30}
\definecolor{captionGainNeutral}{HTML}{6A7075}
\definecolor{captionError}{HTML}{BE4141}
\definecolor{captionUncertaintyLow}{HTML}{142396}
\definecolor{captionUncertaintyHigh}{HTML}{D21919}
\definecolor{cvprblue}{rgb}{0.21,0.49,0.74}
\usepackage[breaklinks,colorlinks,allcolors=cvprblue]{hyperref}
\usepackage[resetlabels]{multibib}
\newcites{supp}{References}
\usepackage{stfloats}
\usepackage[accsupp]{axessibility}
\hypersetup{pdftitle={AdaSeg: Image-Adaptive Configuration for Open-Vocabulary Remote Sensing Semantic Segmentation},
  pdfauthor={Shijun Meng, Shiwei Zhang, Yuhang He, Wei Ke}}

\def\confName{CVPR}
\def\confYear{2027}
\title{AdaSeg: Image-Adaptive Configuration for Open-Vocabulary Remote Sensing Semantic Segmentation}
\author{%
  \textbf{Shijun Meng}\textsuperscript{1,*}\quad
  \textbf{Shiwei Zhang}\textsuperscript{2,*}\quad
  \textbf{Yuhang He}\textsuperscript{3,\textdagger}\quad
  \textbf{Wei Ke}\textsuperscript{1,\textdagger}\\[2pt]
  {\textsuperscript{1}Xi'an Jiaotong University, School of Software Engineering}\\
  {\textsuperscript{2}University of Electronic Science and Technology of China, Institute of Intelligent Computing}\\
  {\textsuperscript{3}Xi'an Jiaotong University, College of Artificial Intelligence}
}

\makeatletter
\patchcmd{\@maketitle}{\vskip .375in}{\vskip .125in}{}%
  {\PackageError{manuscript}{Could not adjust title top spacing}{Check the title macro.}}
\patchcmd{\@maketitle}{\vskip .5em}{\vskip 0pt}{}%
  {\PackageError{manuscript}{Could not adjust author bottom spacing}{Check the title macro.}}
\patchcmd{\@maketitle}{\vspace*{24pt}}{\vspace*{6pt}}{}%
  {\PackageError{manuscript}{Could not adjust author spacing}{Check the title macro.}}
\patchcmd{\@maketitle}{\vspace*{12pt}}{\vspace*{0pt}}{}%
  {\PackageError{manuscript}{Could not adjust title-block spacing}{Check the title macro.}}
\newcommand{\firstpagefigure}{%
  \begingroup
  \begin{minipage}{\textwidth}
    \centering
    \captionsetup{type=figure,position=bottom,skip=3pt}%
    \includegraphics[width=\textwidth]{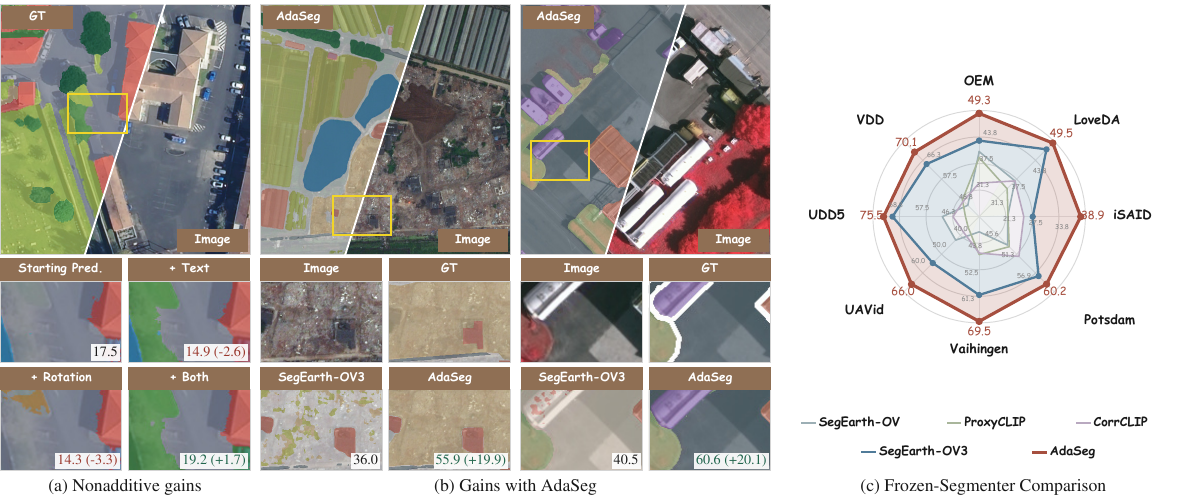}
    \par
    \caption{\textbf{Operator Composition.} (a) Nonadditive effects of operator combinations on segmentation quality. (b) Qualitative comparisons illustrating segmentation improvements with AdaSeg. (c) Radar chart comparing the segmentation performance of AdaSeg and four other frozen-segmenter methods across eight remote sensing datasets.}
 \label{fig:firstpage-motivation}
  \end{minipage}\par
  \endgroup
  \vspace{8pt}%
}
\patchcmd{\@maketitle}{\end{center}}{\end{center}\firstpagefigure}{}{\PackageError{manuscript}{Could not insert first-page figure}{Check the title macro.}}
\makeatother

\begin{document}
\raggedbottom
\maketitle
\begingroup
  \renewcommand{\thefootnote}{}%
  \begin{NoHyper}
  \footnotetext{\textsuperscript{*}Equal contribution.\quad
    \textsuperscript{\textdagger}Corresponding authors.}%
  \end{NoHyper}
\endgroup
\begin{abstract}
Open-vocabulary remote sensing semantic segmentation (OVRSSS) supports diverse Earth observation applications by accommodating flexible semantic vocabularies. Recent approaches introduce inference-time operators (e.g., text expansion and rotation) for semantic recognition and spatial localization. However, their dataset-level gains can conceal regressions on individual images, while nonadditive operator interactions further complicate configuration selection. To address this issue, we initially construct FLAIR-Operator, a dataset that links image evidence and starting configurations to measured operator responses and source-supported sampling distributions. Then, we propose AdaSeg, which uses this dataset to train a vision-language policy for image-adaptive configuration selection through supervised fine-tuning followed by reinforcement learning. Conditioned on image content and prediction feedback, the policy generates an operator configuration that a frozen segmenter executes to produce the final prediction. Experiments on eight remote sensing benchmarks show that AdaSeg achieves an average mIoU of 59.9\%, exceeding the strongest frozen-segmenter baseline by 3.8 percentage points. Our code and FLAIR-Operator dataset will be released.
\end{abstract}

\section{Introduction}
\label{sec:intro}

Open-vocabulary Semantic Segmentation (OVSS) aims to assign pixel-level labels to categories specified in natural language, thereby extending semantic segmentation beyond a predefined training vocabulary. Building on pretrained vision-language models~\cite{DBLP:conf/icml/RadfordKHRGASAM21,DBLP:conf/icml/JiaYXCPPLSLD21,DBLP:conf/iccv/ZhaiM0B23}, OVSS methods transfer image-text alignment to dense prediction~\cite{DBLP:conf/iclr/LiWBKR22,DBLP:conf/cvpr/XuZWHB23,DBLP:conf/cvpr/XieC0KP24,DBLP:conf/cvpr/ChoSHASK24}. Further advances in attention and feature refinement improve local discrimination and spatial consistency across diverse categories and scenes~\cite{DBLP:conf/eccv/WangMY24,DBLP:conf/eccv/LanCKWFZ24,DBLP:conf/cvpr/BousselhamPFK24,DBLP:conf/eccv/LanCKWFZ24a,DBLP:conf/iccv/ZhangLT25}. These advances have also encouraged the extension of OVSS to specialized domains, particularly Open-vocabulary Remote Sensing Semantic Segmentation (OVRSSS).

OVRSSS methods extend OVSS to remote sensing imagery, enabling pixel-level interpretation of land-cover categories and objects across diverse Earth observation scenarios~\cite{DBLP:conf/aaai/YeZZ25,DBLP:conf/cvpr/LiLCBZM025,DBLP:conf/cvpr/ZhuLJLWZRWC0LZ25,DBLP:journals/corr/abs-2512-08730}. Although text-defined categories offer a flexible interface for these interpretation needs, transferring segmentation capabilities to overhead imagery remains challenging. Objects vary substantially in scale and orientation, small instances often appear densely distributed, and visually similar categories must be distinguished within complex spatial layouts~\cite{DBLP:conf/cvpr/DuttaVGRB25,DBLP:conf/cvpr/LiLCBZM025,DBLP:journals/corr/abs-2512-08730,DBLP:conf/iccv/HuangHQW25}. Existing approaches address these challenges through a range of inference-time operations, including textual prompting and semantic adaptation~\cite{DBLP:conf/cvpr/DuttaVGRB25,DBLP:conf/iccv/HuangHQW25}, geometric and spatial refinement~\cite{DBLP:conf/cvpr/DuttaVGRB25,DBLP:conf/cvpr/LiLCBZM025}, and prediction fusion or category filtering~\cite{DBLP:journals/corr/abs-2512-08730}. However, dataset-level improvements from these operations do not guarantee benefits for every image: aggregate gains can conceal regressions on individual images. Our pilot study on FLAIR~\cite{DBLP:conf/nips/GarioudGLWVPGW23} illustrates this issue. As shown in \cref{fig:firstpage-motivation}(a), adding text descriptions or a rotated view individually reduces the mIoU of the default prediction for a FLAIR image by 2.64 and 3.26 percentage points, respectively. Moreover, interactions between operations make their combined effects nonadditive. In the same example, applying both operations together increases mIoU by 1.68 percentage points, even though each operation alone is harmful. Evaluating operations in isolation therefore cannot fully predict their effects within a combined configuration. \cref{sec:main-observation} further quantifies both the variation in operation gains across images and the nonadditive effects of operation combinations.

To address these challenges, we focus on inference-time operations and define an \emph{operator} as a configurable inference-time operation that changes how a frozen segmenter is queried or how its outputs are aggregated and refined, without updating model parameters. For example, head-independent text expansion adds descriptions to a separate set for each prediction head. A fusion operator specifies how head responses are combined. A refinement operator adds a geometric view, a local crop, a RemoteCLIP class prior~\cite{DBLP:journals/tgrs/LiuCGZZYFZ24}, or DINOv3 affinity smoothing~\cite{DBLP:journals/tmlr/SimeoniVSBOJKSYRMHWWDMS26}.
An operator \emph{configuration} specifies the selected operators and their settings, including activation and strength for refinement operators. Subsequently, we propose \textbf{\method}, an open-vocabulary remote sensing semantic segmentation framework that generates image-adaptive operator configurations that account for both image-dependent benefits and operator interactions.
We first construct FLAIR-Operator by executing candidate operators on labelled source images and measuring their effects on segmentation quality. The dataset associates image evidence and starting configurations with operator responses and source-supported sampling distributions, enabling a vision-language policy to learn configuration generation.
The policy combines the RGB image, an initial segmentation, a continuous uncertainty map and marked candidate crop regions, together with class definitions and the starting configuration to select text descriptions, prediction fusion settings, and refinement operators with their activation choices and strengths.
The policy is trained through supervised fine-tuning followed by reinforcement learning, using separate feedback for description additions and refinement decisions.
Executed segmentation gains provide feedback on complete configurations, allowing refinement decisions to be evaluated in the context of other selected choices.
At inference, a single policy call generates a complete configuration, which the frozen segmentation model executes to produce the final prediction.

Our main contributions are summarized as follows:
\begin{itemize}
    \item  We introduce FLAIR-Operator, a dataset that associates remote sensing images and starting operator configurations with measured operator responses and source-supported sampling distributions, enabling a policy to learn to generate image-adaptive operator configurations.

    \item  We propose AdaSeg, a framework for open-vocabulary remote sensing semantic segmentation that generates image-adaptive operator configurations. Conditioned on image content and the segmentation model’s prediction feedback, the framework automatically selects text descriptions, prediction fusion settings, and refinement operators for inference.

    \item We evaluate AdaSeg on eight remote sensing segmentation benchmarks spanning satellite, aerial, and UAV imagery. The results show that our method achieves an average mIoU of 59.9\%, exceeding all previous approaches.
\end{itemize}

\section{Related Work}
\label{sec:related-work}

\boldheading{Open-Vocabulary Segmentation.} Open-vocabulary segmentation adapts visual--language representations through learned alignment and mask recognition~\cite{DBLP:conf/iclr/LiWBKR22,DBLP:conf/cvpr/XuZWHB23,DBLP:conf/cvpr/XieC0KP24,DBLP:conf/cvpr/ChoSHASK24}, training-free attention and feature refinement~\cite{DBLP:conf/eccv/ZhouLD22,DBLP:conf/eccv/WangMY24,DBLP:conf/cvpr/BousselhamPFK24,DBLP:conf/eccv/LanCKWFZ24}, or spatial guidance from auxiliary visual models~\cite{DBLP:conf/eccv/LanCKWFZ24a,DBLP:conf/iccv/ZhangLT25}. ReAttnCLIP redefines attention in CLIP for training-free remote-sensing segmentation~\cite{Niu_2026_CVPR}. These approaches recover spatial detail while retaining the semantic flexibility of pretrained representations. Remote-sensing methods further incorporate domain supervision, specialist features, and upsampling~\cite{DBLP:conf/aaai/YeZZ25,DBLP:conf/cvpr/ZhuLJLWZRWC0LZ25,DBLP:conf/cvpr/LiLCBZM025}. AerOSeg combines rotation-aware correlations with SAM-guided refinement~\cite{DBLP:conf/cvpr/DuttaVGRB25}, while SCORE uses regional and global context for remote-sensing instance segmentation~\cite{DBLP:conf/iccv/HuangHQW25}.

SAM~3 supports concept segmentation from text and image exemplars~\cite{DBLP:conf/iclr/CarionGHDHCRAKH26}. SegEarth-OV3 adapts it to remote-sensing semantic segmentation by fusing semantic and instance predictions and suppressing absent categories through presence scores~\cite{DBLP:journals/corr/abs-2512-08730}. ActiveSAM uses an initial presence-screening stage to filter the vocabulary before segmentation with frozen SAM~3~\cite{tien2026activesam}. AdaSeg builds on these capabilities to configure a frozen segmenter for individual images. Because operator gains depend on image content and operator interactions, we jointly select descriptions and execution settings from visual and prediction evidence.

\boldheading{Active Visual Perception.} Language-based models organize visual computation through program generation and tool composition~\cite{DBLP:conf/cvpr/GuptaK23,DBLP:conf/iccv/SurisMV23,DBLP:conf/eccv/LiuCLZLRZYSZZGL24}. Reinforcement learning further enables task-driven region selection and visual interaction~\cite{zhu2026activeo3empoweringmllmsactive,DBLP:journals/corr/abs-2505-14246,DBLP:conf/iclr/ZhengYHZXYSY26,DBLP:conf/nips/SuWRLC25}. ACTIVE-o3 learns region-selection strategies through GRPO without explicit region-selection supervision~\cite{zhu2026activeo3empoweringmllmsactive}; Visual-ARFT, DeepEyes, and Pixel Reasoner learn visual interactions for multimodal reasoning~\cite{DBLP:journals/corr/abs-2505-14246,DBLP:conf/iclr/ZhengYHZXYSY26,DBLP:conf/nips/SuWRLC25}. AdaSeg instead learns segmentation configurations, generating descriptions and refinement settings in one policy call. The policy uses RGB, initial predictions, uncertainty, and crop evidence to choose a complete configuration, with feedback evaluating the selected operations together rather than in isolation. FLAIR-Operator provides measured operator responses for supervision, and execution-based rewards connect configuration choices to segmentation quality while the segmenter remains frozen.

\section{Preliminary}
\label{sec:prelim}

Operator selection predicts an operator configuration for a frozen segmentation model.
We define the configuration space and characterize operator responses through per-image mIoU gains and pairwise interactions.
These quantities link image-dependent operator responses to the supervision used in FLAIR-Operator.

\subsection{Configurations and Operators}
\label{sec:characterization}
\label{sec:setup}
Given an image $x$ and target classes $\mathcal C$, a policy selects configuration $a$ for a frozen segmenter $F_\phi$:
\begin{equation}
  \hat y=F_\phi(x;a),\qquad a\sim\pi_\theta(\cdot\mid o_x).
  \label{eq:task}
\end{equation}
The observation $o_x$ contains the image, class definitions, candidate descriptions, starting configuration, and segmenter responses.
Write $a=(D,f,b,\mathbf w)$, where $D$ contains the head-specific description sets, $f$ the fusion settings, and $b$ the background rule.
Head-independent text expansion adds descriptions separately to $D$ for the instance, semantic, and presence heads.
It retains the starting descriptions and target classes $\mathcal C$.
The refinement weights are $\mathbf w=(w_1,\ldots,w_J)$, with $w_j=g_js_j$.
Here $g_j\in\{0,1\}$ enables refinement operator $j$, and $s_j\in\mathcal S^+=\{0.1,0.25,0.5\}$ is its conditional strength.
Source probing, supervision, and policy output use the same three positive strengths.
The ten refinement choices comprise three rotations, two flips, three crop regions, RemoteCLIP, and DINOv3.
An operator execution changes its designated fields while holding the remaining configuration fixed.
This shared configuration space lets the policy coordinate semantic matching and spatial refinement for each image.

\begin{figure}[t]
  \centering
  \includegraphics[width=\columnwidth]{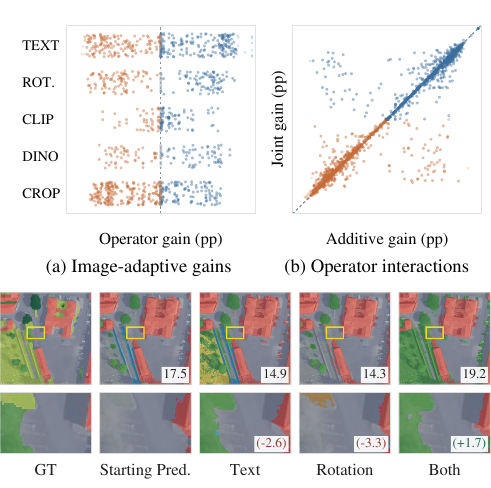}
  \caption{\textbf{Operator Gains and Interactions.} (a) Individual gains. (b) Joint versus summed individual gains. Colors indicate \textcolor{captionGainPositive}{positive}, \textcolor{captionGainNegative}{negative}, and \textcolor{captionGainNeutral}{near-zero} gains. Dashed lines mark zero gain and additive agreement.}
  \label{fig:observation}
  \label{fig:action-interactions}
\vspace{-2pt}
\end{figure}

\begin{figure*}[t]
 \centering
 \includegraphics[width=\textwidth]{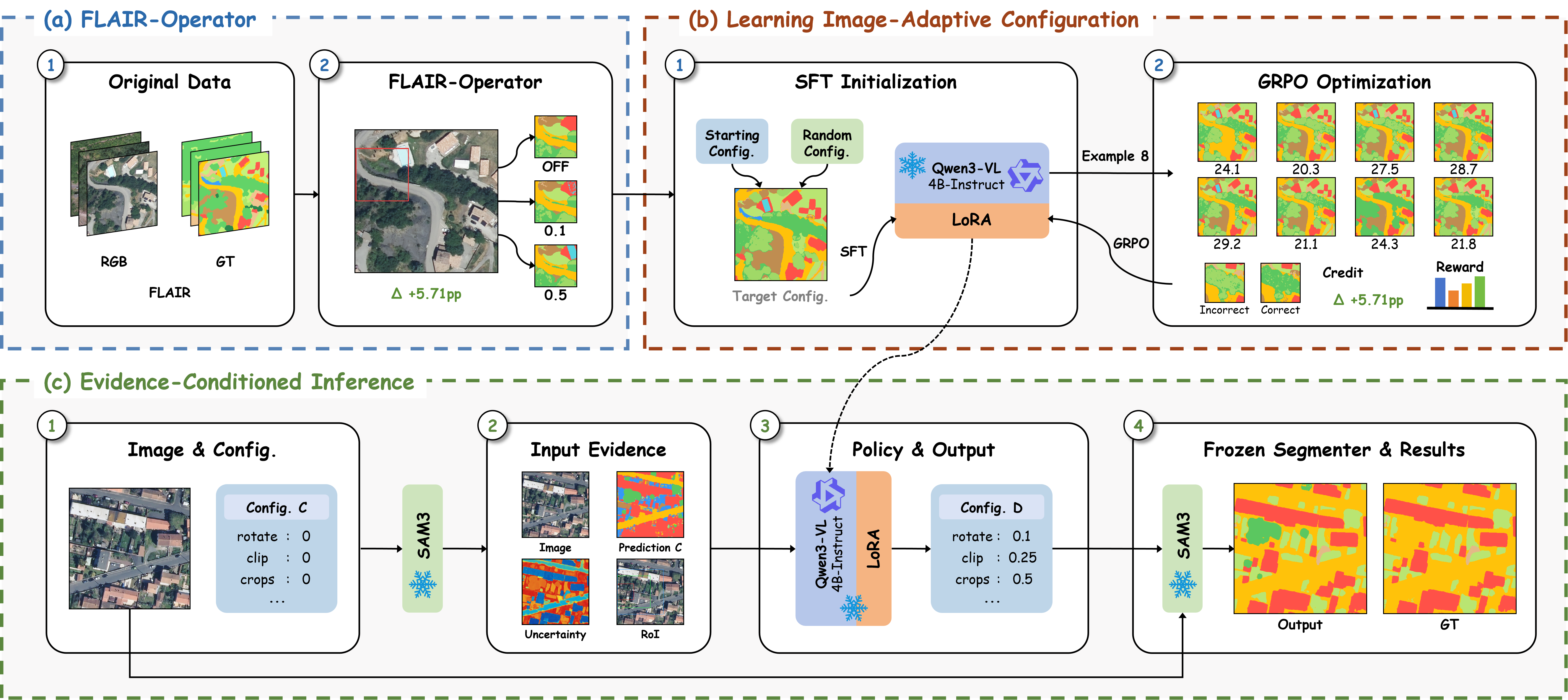}
 \par\vspace{0.008982416478\textwidth}
 \caption{\textbf{\method Overview.} (a) FLAIR-Operator construction from source images and measured operator responses. (b) Image-adaptive configuration learning through supervised initialization and reward-based optimization. (c) Evidence-conditioned inference, where the policy selects a complete configuration from visual evidence and frozen SAM~3 executes it to produce the final segmentation.}
 \label{fig:overview}
\vspace{-2pt}
\end{figure*}

\subsection{Gains and Operator Interactions}
\label{sec:effects}
\label{sec:interactions}
\label{sec:observation}
For image $i$, define the gain over starting configuration $a_i^0$ as
\begin{equation}
 \Delta_i(a)=m_i(a)-m_i(a_i^0),
 \label{eq:gain}
\end{equation}
where $m_i$ is per-image mIoU and gains are reported in percentage points (pp).
The reference configuration fixes the comparison for each image, so the gain measures the change caused by the selected configuration.
For operators $j$ and $k$, let $\Delta_{i,j}$ and $\Delta_{i,k}$ be their individual gains, and $\Delta_{i,jk}$ their joint gain.
The operator interaction is the departure of the joint gain from additivity:
\begin{equation}
 J_{i,jk}=\Delta_{i,jk}-\Delta_{i,j}-\Delta_{i,k}.
 \label{eq:residual}
\end{equation}
Positive and negative residuals indicate joint gains above and below the additive expectation.
The sign of $\Delta_{i,jk}$ determines whether the combined configuration improves the starting prediction.
All comparisons keep the image, target classes, and model parameters fixed.

Source probes use the same paired comparison, with the reference differing only in the tested description or refinement setting.
Removing a description or disabling an operator within a joint configuration yields its conditional mIoU gain.
Each record retains the reference settings, allowing the same operator to be compared across distinct surrounding configurations.
The conditional gain evaluates whether an operator benefits a given combination, providing context for jointly selecting descriptions and refinement settings.

\subsection{Main Observations}
\label{sec:main-observation}
Measured operator responses on the FLAIR pilot exhibit image dependence and nonadditive operator interactions (\cref{fig:observation}).
First, individual gains cross zero across images (\cref{fig:observation}(a)): operators (e.g., text expansion and rotation) improve some predictions and degrade others.
Second, pairwise residuals occur on both sides of zero (\cref{fig:observation}(b)): points above and below the diagonal show that joint gains can exceed or fall below the additive expectation.
The residual measures operator interaction, while the joint gain measures improvement over the starting prediction.
These properties motivate image-adaptive selection and feedback that accounts for the surrounding configuration.

\section{Methodology}
\label{sec:method}

\method comprises source-data construction, image-adaptive configuration learning, and evidence-conditioned inference (\cref{fig:overview}).
We first execute candidate operators on labelled FLAIR images and measure their quality changes (\cref{fig:overview}(a)).
FLAIR-Operator stores these responses with image evidence, starting configurations, and sampling distributions (\cref{sec:teacher}).
A vision-language policy then learns to generate complete configurations through supervised initialization and policy optimization (\cref{fig:overview}(b)).
During optimization, executed configurations yield rewards for descriptions, refinement operators, and output format (\cref{sec:training}).
At inference, the starting prediction supplies image-adaptive evidence for one policy call (\cref{fig:overview}(c)).
The policy selects a configuration, which frozen SAM~3 executes to produce the final segmentation (\cref{sec:policy}).

\subsection{FLAIR-Operator}
\label{sec:teacher}
We build FLAIR-Operator from labelled FLAIR images~\cite{DBLP:conf/nips/GarioudGLWVPGW23} and frozen SAM~3 executions~\cite{DBLP:conf/iclr/CarionGHDHCRAKH26}.
Each record links image evidence and a starting configuration to candidate descriptions, refinement settings, and measured operator responses, including per-image mIoU gains.

\begin{figure}[!t]
  \centering
  \includegraphics[width=\columnwidth]{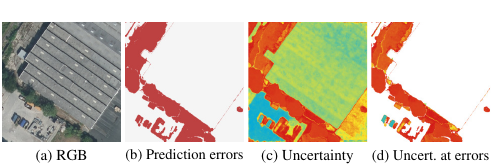}
  \par\vspace{-2pt}
  \caption{\textbf{Uncertainty and Errors.} (a) RGB. (b) \textcolor{captionError}{Errors}. (c) Uncertainty. (d) Uncertainty at error pixels. The shared scale runs from \textcolor{captionUncertaintyLow}{low uncertainty} (0) to \textcolor{captionUncertaintyHigh}{high uncertainty} (1); white in (d) marks non-error pixels.}
  \label{fig:response}
  \par\vspace{0pt}
  \captionsetup{type=figure}
  \includegraphics[width=\columnwidth]{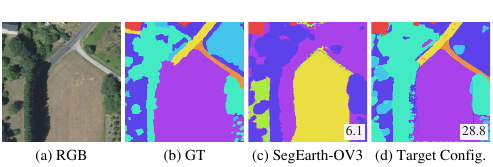}
  \par\vspace{-2pt}
  \caption{\textbf{FLAIR-Operator Example.} (a) RGB. (b) Ground truth. (c) SegEarth-OV3. (d) Target Config. Predictions share the ground-truth class palette. Scores are full-image mIoU (\%) over all 13 classes.}
  \label{fig:teacher}
\vspace{-2pt}
\end{figure}

\paragraph{Image Evidence.}
The starting configuration $a_i^0$ provides four evidence images: RGB, initial segmentation, continuous uncertainty, and RGB marked with three crop regions.
Uncertainty uses the starting configuration's original-view class scores before augmentation:
\begin{equation}
  U_i(p)=1-\clip\!\left(F^0_{(1)}(p)-F^0_{(2)}(p),0,1\right),
  \label{eq:uncertainty}
\end{equation}
where $F^0_{(1)}$ and $F^0_{(2)}$ are the two highest scores; larger uncertainty indicates closer competing classes (\cref{fig:response}).
Crops selected from score ambiguity under fixed reference descriptions remain fixed across configurations.
Class definitions, the starting configuration, and candidate descriptions complete the policy input.

\paragraph{Source Responses and Sampling.}
Head-independent text expansion adds descriptions per class and head, preserving starting descriptions and target classes.
Each probe tests one description addition or refinement strength with other settings fixed.
Refinements without positive individual support are also tested in sampled joint configurations.
Measured responses define bounded support, setting nonpositive responses to zero.
Within FLAIR-Operator, sampling support is obtained by distance-weighted aggregation of operator responses from similar source images, excluding the current image.

Descriptions are sampled independently from their supported inclusion probabilities.
For each refinement operator, the strongest supported setting determines activation support, while normalized support determines its conditional strength distribution.
A shared exploration coefficient $\eta$ mixes each supported distribution with uniform exploration.
Sampled targets preserve starting descriptions and fusion fields.
All-zero strength support defaults to uniform conditional sampling.
\cref{fig:teacher} illustrates a target configuration against SegEarth-OV3.

\FloatBarrier
\subsection{Learning Image-Adaptive Configurations}
\label{sec:training}
\label{sec:sft}
\label{sec:credit}
\label{sec:description-reward}
\label{sec:grpo}
We use Qwen3-VL-4B~\cite{bai2025qwen3vltechnicalreport} with LoRA adapters~\cite{DBLP:conf/iclr/HuSWALWWC22}, keeping the base model and segmentation executor frozen.
The language output independently expands each class's head-specific description sets and specifies fusion fields.
Each refinement operator's readout combines a learned embedding with the final input hidden state and attends to input visual tokens.
Separate categorical outputs predict activation and conditional strength, yielding $w_j=g_js_j$ and a complete JSON configuration.

\paragraph{Supervised Initialization.}
Each training presentation draws a fresh target from the source-supported distributions, including configurations with negative executed gains.
The task loss averages normalized losses for descriptions, refinement operators, and fusion, while syntax and field names receive a separate format loss:
\begin{equation*}
  \mathcal L_{\mathrm{SFT}}=\mathcal L_{\mathrm{task}}+\mathcal L_{\mathrm{fmt}}.
\end{equation*}
Refinement operators are supervised through the joint activation--strength decision; disabled operators contribute only their activation loss.

\paragraph{Reward Design.}
For each image, the policy samples a group of complete configurations and receives separate feedback for descriptions and refinement operators.
For executable configurations, define $\gamma_k=\mathbb I[\Delta_k\ge-\tau]$, where $\Delta_k=m_i(a_{ik})-m_i(a_i^0)$ is the mIoU gain and $\tau$ the regression tolerance.

Description credit measures the marginal coverage of source-supported candidate descriptions.
For head $h$, let $\mathcal R_h$ contain the candidate descriptions, $p_q$ their inclusion probabilities, and $S_{kh}$ the selected additions.
With target mass $T_h=\sum_{q\in\mathcal R_h}p_q^2$, coverage is
\begin{equation}
  C_h(S_{kh})=\min\!\left(1,
  \frac{\sum_{q\in S_{kh}\cap\mathcal R_h}p_q}{T_h}\right).
  \label{eq:coverage}
\end{equation}
An added candidate description receives marginal coverage credit:
\begin{equation}
  B_{kq}=C_h(S_{kh})-C_h(S_{kh}\setminus\{q\}).
  \label{eq:reward}
\end{equation}
Only added candidate descriptions earn coverage credit.
Array-closing feedback accounts for uncovered support and missing starting descriptions, while duplicate additions receive a fixed penalty.

Refinement rewards use $\alpha$ to balance bounded mIoU gain $u_k$ against operator-specific agreement $R_j^T$ with the source-supported distribution.
Rewards are centered and clipped within each image--operator group to obtain $B_{kj}$.
Fusion and background fields use the mean refinement credit before gating.
For each description, refinement, fusion, or background decision $d$, the final advantage is
\begin{equation}
  \begin{split}
  r_{kj}&=\alpha u_k+(1-\alpha)R_j^T(w_{kj}),\\
  A_{kd}&=\min(B_{kd},0)+\gamma_k\max(B_{kd},0).
  \end{split}
  \label{eq:credit}
\end{equation}
The shared gate suppresses positive task advantages for configurations whose mIoU regression exceeds the tolerance.
Executed mIoU feedback uses FLAIR training labels.
Output validity uses an independent format objective, including for unexecutable outputs.

\begin{table*}[!t]
 \centering
 \fontsize{8}{9.5}\selectfont
 \setlength{\tabcolsep}{3pt}
 \renewcommand{\arraystretch}{0.90}
 \caption{\textbf{Quantitative comparison on eight remote sensing OVSS benchmarks.} \textcolor{red}{Best} and \textcolor{blue}{second-best} frozen-segmenter scores are highlighted. All values are mIoU (\%), and Avg. is computed over the reported entries.}
 \label{tab:main-results}
 \begin{tabular*}{\textwidth}{@{\extracolsep{\fill}}l*{9}{c}@{}}
 \toprule
 \multirow{2}{*}{\textbf{Method}} & \multicolumn{8}{c}{\textbf{Dataset}} & \multicolumn{1}{l}{\multirow{2}{*}{\textbf{Avg.}}}\\
 \cmidrule(lr){2-9}
 & \textbf{OEM} & \textbf{LoveDA} & \textbf{iSAID} & \textbf{Potsdam} & \textbf{Vaihingen} & \textbf{UAVid} & \textbf{UDD5} & \textbf{VDD} & \\
 \midrule
 \rowcolor{gray!15} \multicolumn{10}{c}{\textit{Trained-Segmenter Methods}}\\
 \midrule
 Cat-Seg\methodsource{CVPR'24}~\cite{DBLP:conf/cvpr/ChoSHASK24} & - & 28.6 & 53.3 & 35.8 & 42.3 & 25.7 & 40.2 & 39.1 & -\\
 RSKT-Seg\methodsource{AAAI'26}~\cite{DBLP:conf/aaai/LiDZZSG26} & - & 33.2 & 54.3 & 38.4 & 42.7 & 25.7 & 42.1 & 39.7 & -\\
 GSNet\methodsource{AAAI'25}~\cite{DBLP:conf/aaai/YeZZ25} & - & 32.5 & 53.7 & 37.9 & 44.1 & 24.2 & 40.9 & 37.3 & -\\
 SkySense-O\methodsource{CVPR'25}~\cite{DBLP:conf/cvpr/ZhuLJLWZRWC0LZ25} & 40.8 & 38.3 & 43.9 & 54.1 & 51.6 & - & - & - & -\\
 \midrule
 \rowcolor{gray!15} \multicolumn{10}{c}{\textit{Frozen-Segmenter Methods}}\\
 \midrule
 GEM\methodsource{CVPR'24}~\cite{DBLP:conf/cvpr/BousselhamPFK24} & 33.9 & 31.6 & 17.7 & 39.1 & 36.4 & 33.4 & 41.2 & 39.5 & 34.1\\
 ClearCLIP\methodsource{ECCV'24}~\cite{DBLP:conf/eccv/LanCKWFZ24} & 31.0 & 32.4 & 18.2 & 42.0 & 36.2 & 36.2 & 41.8 & 39.3 & 34.6\\
 ProxyCLIP\methodsource{ECCV'24}~\cite{DBLP:conf/eccv/LanCKWFZ24a} & 38.9 & 34.3 & 21.8 & 49.0 & 47.5 & 35.8 & 40.8 & 47.8 & 39.5\\
 SegEarth-OV\methodsource{CVPR'25}~\cite{DBLP:conf/cvpr/LiLCBZM025} & 40.3 & 36.9 & 21.7 & 48.5 & 40.0 & 42.5 & 50.6 & 45.3 & 40.7\\
 CorrCLIP\methodsource{ICCV'25}~\cite{DBLP:conf/iccv/ZhangLT25} & 32.9 & 36.9 & 25.5 & 51.9 & 47.0 & 38.3 & 46.1 & 47.3 & 40.7\\
 ReAttnCLIP\methodsource{CVPR'26}~\cite{Niu_2026_CVPR} & 41.1 & 37.0 & 23.2 & 48.7 & 29.9 & 44.0 & 53.7 & 49.7 & 40.9\\
 ActiveSAM\textsuperscript{*}\methodsource{arXiv'26}~\cite{tien2026activesam} & 40.7 & 41.1 & 28.3 & 40.6 & 45.1 & 53.3 & 72.1 & 62.2 & 47.9\\
 SegEarth-OV3\methodsource{arXiv'25}~\cite{DBLP:journals/corr/abs-2512-08730} & 42.9 & \textcolor{blue}{47.4} & 27.6 & 57.8 & 60.8 & 54.7 & 71.7 & 64.5 & 53.4\\
 ProC-SAM3\methodsource{GRSL'26}~\cite{song2026procsam3} & \textcolor{blue}{48.9} & 46.1 & \textcolor{blue}{29.5} & \textcolor{blue}{58.8} & \textcolor{blue}{64.5} & \textcolor{blue}{60.0} & \textcolor{blue}{73.4} & \textcolor{blue}{67.5} & \textcolor{blue}{56.1}\\
 \textbf{Ours} & \textcolor{red}{49.3} & \textcolor{red}{49.5} & \textcolor{red}{38.9} & \textcolor{red}{60.2} & \textcolor{red}{69.5} & \textcolor{red}{66.0} & \textcolor{red}{75.5} & \textcolor{red}{70.1} & \textcolor{red}{59.9}\\
 \bottomrule
 \end{tabular*}
\end{table*}

\paragraph{Policy Optimization.}
We refine the supervised policy with group relative policy optimization (GRPO)~\cite{DBLP:journals/corr/abs-2402-03300}:
\begin{equation}
  \mathcal L=\mathcal L_{\mathrm{GRPO}}+\mathcal L_{\mathrm{fmt}}^{\mathrm{RL}}+\lambda_{\mathrm{rep}}\mathcal L_{\mathrm{SFT}}.
  \label{eq:objective}
\end{equation}
Advantages for descriptions and refinement operators enter clipped probability-ratio objectives, normalized separately and weighted equally.
All task decisions use the gated advantages $A_{kd}$; refinement updates use joint activation--strength probabilities.

KL regularization constrains deviation from the fixed supervised policy, while $\lambda_{\mathrm{rep}}$ weights source-supported SFT replay loss to retain starting descriptions and output syntax.
Each round freezes the rollout policy and evaluates current-policy probabilities at the same temperature.

\subsection{Evidence-Conditioned Inference}
\label{sec:evidence}
\label{sec:policy}
Frozen SAM~3 executes $a_i^0$ to obtain image evidence (\cref{sec:teacher}).
The policy selects a configuration for the image or evaluation window in one call (\cref{fig:overview}(c)), using the visual evidence and target class definitions.
The starting prediction provides an executor-specific reference, while uncertainty and crop views guide semantic expansion and spatial refinement.

Within each class and head, maximum responses over selected descriptions are fused across instance and semantic heads and weighted by class presence.
Alternative descriptions let each class draw on complementary semantic matches.
The executor combines aligned geometric and crop predictions with these original-view scores, then applies enabled RemoteCLIP class priors and DINOv3 affinity refinement.
Alignment places these responses in the original image coordinates for spatially consistent fusion.
Geometric views cover the image; crops affect eligible pixels within their regions.
A binary mask recomputed from these scores gates local updates, separately from input uncertainty.
The background rule converts refined scores into class labels.
Unmasked pixels retain their scores during local refinement.
Selected descriptions preserve the identity of each target class.

\section{Experiments}
\label{sec:exp}

\begingroup
 \renewcommand{\thefootnote}{*}%
 \footnotetext[1]{ActiveSAM results were reproduced by ProC-SAM3~\cite{song2026procsam3}.}%
\endgroup

In this section, we evaluate AdaSeg for open-vocabulary remote sensing semantic segmentation.
We first detail the implementation in \cref{sec:datasets}, then present quantitative comparisons on eight benchmarks in \cref{sec:main-results}.
Ablation studies analyze configuration selection and execution cost in \cref{sec:ablation}, and qualitative comparisons are provided in \cref{sec:qualitative-results}.
More details are in the supplementary material.

\subsection{Implementation}
\label{sec:datasets}
We construct FLAIR-Operator by executing candidate text expansions and refinement operators on labelled FLAIR images~\cite{DBLP:conf/nips/GarioudGLWVPGW23} with frozen SAM~3~\cite{DBLP:conf/iclr/CarionGHDHCRAKH26}.
We use source annotations to measure per-image mIoU changes relative to a starting configuration.
Refinement operators lacking positive individual support are also tested within joint configurations to capture operator interactions.
Responses define source-supported sampling distributions for configuration targets in supervised fine-tuning (SFT).
Individual and conditional probes provide complementary evidence about an operator's utility across configurations.
Each sampled target retains its associated image evidence.

We use Qwen3-VL-4B~\cite{bai2025qwen3vltechnicalreport} with LoRA adapters~\cite{DBLP:conf/iclr/HuSWALWWC22}, keeping the base model and segmenter frozen.
GRPO refines the policy initialized by SFT using configuration execution feedback.
Training uses $4\times$A100~GPUs.
We start from the baselines' configurations and build candidate descriptions using common synonyms and text expansions.
A single policy call generates an image-adaptive configuration for execution by the frozen segmenter.
This setup learns how to combine the executor's existing capabilities through configuration selection.

\begin{table*}[!t]
 \centering
 \begin{minipage}[t]{\columnwidth}
 \captionsetup{type=table}
 \vspace{0pt}
 \centering
 \fontsize{8}{9.5}\selectfont
 \setlength{\tabcolsep}{3pt}
 \caption{\textbf{Ablation of policy training.} VLLM denotes Qwen3-VL-4B; $\checkmark$ marks enabled components. The \textcolor{red}{best} and \textcolor{blue}{second-best} scores are highlighted. All values are mIoU (\%); image annotations show per-image scores.}
 \label{tab:training-results}
 \begin{tabular*}{\columnwidth}{@{\extracolsep{\fill}}*{3}{c}*{4}{l}@{}}
 \toprule
 \textbf{VLLM} & \textbf{SFT} & \textbf{GRPO} & \textbf{LoveDA} & \textbf{OEM} & \textbf{UDD5} & \textbf{Vaihingen}\\
 \midrule
  &  &  & \textcolor{blue}{47.4} & 42.9 & 71.7 & \textcolor{blue}{60.8}\\
 $\checkmark$ &  &  & \textcolor{blue}{47.4}\textsuperscript{*} & 42.9\textsuperscript{*} & 71.7\textsuperscript{*} & \textcolor{blue}{60.8}\textsuperscript{*}\\
 $\checkmark$ & $\checkmark$ &  & 46.8 & \textcolor{blue}{46.5} & \textcolor{blue}{73.9} & 36.1\\
 $\checkmark$ & $\checkmark$ & $\checkmark$ & \textcolor{red}{49.5} & \textcolor{red}{49.3} & \textcolor{red}{75.5} & \textcolor{red}{69.5}\\
 \bottomrule
 \end{tabular*}
 \par\vspace{2pt}
 \noindent\makebox[\columnwidth][l]{$\ast$: Output exactly copies the starting configuration.}
 \par\vspace{4pt}
 \noindent\includegraphics[width=\columnwidth]{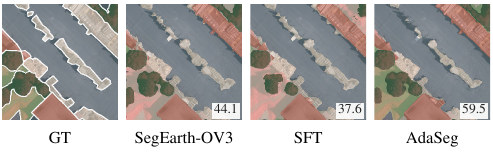}
\end{minipage}\hfill
 \begin{minipage}[t]{\columnwidth}
 \captionsetup{type=table}
 \vspace{0pt}
 \centering
 \fontsize{8}{9.5}\selectfont
 \setlength{\tabcolsep}{3pt}
 \caption{\textbf{Ablation of configuration components.} Desc.: head-independent text expansion, selecting class descriptions separately for each prediction head. Exec.: image-adaptive fusion settings and refinement operators. $\checkmark$ marks enabled components. The \textcolor{red}{best} and \textcolor{blue}{second-best} scores are highlighted. All variants keep the executor frozen and preserve target classes. All scores are mIoU (\%).}
 \label{tab:component-results}
 \begin{tabular*}{\columnwidth}{@{\extracolsep{\fill}}*{6}{c}@{}}
 \toprule
 \textbf{Desc.} & \textbf{Exec.} & \textbf{LoveDA} & \textbf{OEM} & \textbf{UDD5} & \textbf{Vaihingen}\\
 \midrule
  &  & 47.4 & 42.9 & \textcolor{blue}{71.7} & 60.8\\
 $\checkmark$ &  & 48.2 & \textcolor{blue}{48.4} & 68.4 & \textcolor{blue}{67.7}\\
  & $\checkmark$ & \textcolor{blue}{48.9} & 44.5 & 71.2 & 60.9\\
 $\checkmark$ & $\checkmark$ & \textcolor{red}{49.5} & \textcolor{red}{49.3} & \textcolor{red}{75.5} & \textcolor{red}{69.5}\\
 \bottomrule
 \end{tabular*}
 \par\vspace{4pt}
 \noindent\includegraphics[width=\columnwidth]{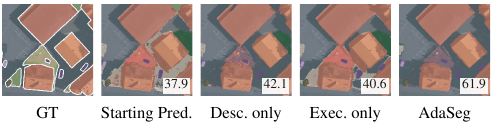}
\end{minipage}
\end{table*}

\begin{table*}[!t]
 \centering
 \begin{minipage}[t]{\columnwidth}
 \captionsetup{type=table}
 \vspace{0pt}
 \centering
 \fontsize{8}{9.5}\selectfont
 \setlength{\tabcolsep}{3pt}
 \caption{\textbf{Ablation of evidence inputs.} ``w/o Evidence'': better RGB-only variant with/without candidate descriptions. Bold: best scores. All values are mIoU (\%).}
 \label{tab:evidence-ablation}
 \begin{tabular*}{\columnwidth}{@{\extracolsep{\fill}}l*{4}{c}@{}}
 \toprule
 \textbf{Setting} & \textbf{LoveDA} & \textbf{OEM} & \textbf{UDD5} & \textbf{Vaihingen}\\
 \midrule
 SegEarth-OV3 & 47.4 & 42.9 & 71.7 & 60.8\\
 AdaSeg w/o Evidence & 42.4 & 47.0 & 67.1 & 50.1\\
 AdaSeg Full & \textbf{49.5} & \textbf{49.3} & \textbf{75.5} & \textbf{69.5}\\
 \bottomrule
 \end{tabular*}
\end{minipage}\hfill
 \begin{minipage}[t]{\columnwidth}
 \captionsetup{type=table}
 \vspace{0pt}
 \centering
 \fontsize{8}{9.5}\selectfont
 \setlength{\tabcolsep}{3pt}
 \caption{\textbf{Per-image changes.} Entries show Non-declined / Declined relative to SegEarth-OV3. Non-declined includes unchanged or higher mIoU; Declined denotes lower mIoU. All values are percentages of images in each dataset.}
 \label{tab:per-image-improvement}
 \begin{tabular*}{\columnwidth}{@{\extracolsep{\fill}}l*{4}{c}@{}}
 \toprule
 \textbf{Method} & \textbf{LoveDA} & \textbf{OEM} & \textbf{UDD5} & \textbf{Vaihingen}\\
 \midrule
 Operator ALL & 36.0 / 64.0 & 3.7 / 96.4 & 22.5 / 77.5 & 1.8 / 98.2\\
 AdaSeg & \textbf{70.5 / 29.5} & \textbf{82.8 / 17.2} & \textbf{80.0 / 20.0} & \textbf{87.7 / 12.3}\\
 \bottomrule
 \end{tabular*}
\end{minipage}
\end{table*}

\subsection{Main Results}
\label{sec:main-results}
We evaluate on OpenEarthMap (OEM)~\cite{DBLP:conf/wacv/XiaYAB23}, LoveDA~\cite{DBLP:conf/nips/WangZMLZ21}, iSAID~\cite{DBLP:conf/cvpr/ZamirAGKSK00XB19}, Potsdam, Vaihingen, UAVid~\cite{LYU2020108}, UDD5~\cite{DBLP:conf/prcv/ChenWLCW18}, and VDD~\cite{DBLP:journals/jvcir/CaiJHGWY25}. Satellite, aerial, and UAV imagery provide varied object scales, layouts, and category vocabularies.
The evaluation covers both broad land-cover regions and densely arranged objects.
Following SegEarth-OV3~\cite{DBLP:journals/corr/abs-2512-08730}, we report dataset-level mIoU and the unweighted average across eight benchmarks.

AdaSeg leads frozen-segmenter methods on all eight benchmarks, averaging 59.9\% mIoU, 3.8 points above ProC-SAM3 and 6.5 above the original SegEarth-OV3 results (\cref{tab:main-results}).
Against ProC-SAM3, gains reach 9.4 points on iSAID, 6.0 on UAVid, and 5.0 on Vaihingen.
Against trained-segmenter methods, AdaSeg leads on seven reported benchmarks; trained-segmenter methods perform better on iSAID.

\subsection{Ablation Studies and Analysis}
\label{sec:ablation}
Ablation studies are conducted on LoveDA, OEM, UDD5, and Vaihingen using the same frozen executor. Unless otherwise specified, all experiments on a given dataset use identical text prompts.

\label{sec:training-results}
\ablationheading{Effect of Policy Training.} The untuned VLLM copies the starting configuration for every image, preserving the starting predictions (\cref{tab:training-results}).
Executable outputs alone therefore do not establish evidence-based operator selection.
SFT improves OEM and UDD5 but reduces LoveDA to 46.8\% and Vaihingen to 36.1\%.
The GRPO-trained policy outperforms SFT on all four datasets, reaching 69.5\% on Vaihingen, 8.7 points above SegEarth-OV3.
Execution feedback (\cref{sec:grpo}) thus aligns configuration choices with segmentation quality beyond source-supported target imitation.

\label{sec:component-results}
\ablationheading{Configuration Components.} Separate and joint ablations evaluate head-independent text expansion and execution settings (\cref{tab:component-results}).
Text expansion contributes more on OEM and Vaihingen, while execution changes contribute more on LoveDA.
Joint selection performs best on every dataset, exceeding the stronger isolated component by 0.6--4.3 points.
On UDD5, the individual variants yield 68.4\% and 71.2\%, below the 71.7\% baseline, yet their combination reaches 75.5\%.
Text expansion changes the responses consumed by fusion and refinement, making its contribution dependent on execution settings.
These operator interactions explain why configuration selection must consider components jointly.

\begin{figure*}[!t]
 \centering
 \includegraphics[width=\textwidth]{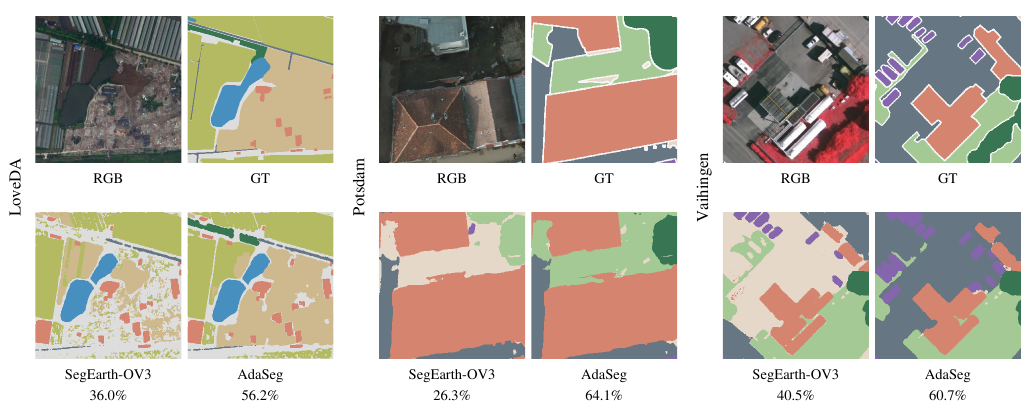}
 \par\vspace{-2pt}
 \caption{\textbf{Qualitative Results.} SegEarth-OV3 and AdaSeg are compared on LoveDA, Potsdam, and Vaihingen. Each example shows the RGB image, ground truth (GT), and predictions from both methods. The comparisons illustrate changes in land-cover consistency and object delineation under image-adaptive configurations. Scores below the predictions report per-image mIoU (\%) over all classes.}
 \label{fig:main-comparison}
\vspace{-2pt}
\end{figure*}

\ablationheading{Effect of Evidence Inputs.} With RGB and the starting configuration retained, the ablation removes prediction, uncertainty, and crop evidence (\cref{tab:evidence-ablation}).
The better variant with or without candidate descriptions is reported.
Restoring full evidence improves mIoU by 2.3--19.4 points, largest on Vaihingen, and exceeds SegEarth-OV3 everywhere rather than only OEM.
The gains indicate that configuration selection benefits from the executor's predictions and ambiguous regions, beyond scene appearance and candidate descriptions.

\begin{samepage}
\ablationheading{Per-Image Improvements.} AdaSeg matches or exceeds SegEarth-OV3's per-image mIoU on 70.5--87.7\% of images, versus 1.8--36.0\% for Operator ALL (\cref{tab:per-image-improvement}).
On OEM and Vaihingen, decline rates are 17.2\% and 12.3\% for AdaSeg, compared with 96.4\% and 98.2\% for Operator ALL.
These proportions complement dataset-level mIoU by showing how frequently configuration selection preserves or improves the starting prediction.
Image-adaptive selection therefore reduces the prevalence of per-image regressions compared with enabling all operators.\par
\end{samepage}

\begin{table}[t]
 \centering
 \fontsize{8}{9.5}\selectfont
 \setlength{\tabcolsep}{1pt}
 \caption{\textbf{Configuration selection and execution cost.} Cost counts executions per image. R@$K$: $K$ sampled configurations; $+$/$-$: \textcolor{captionGainNeutral}{GT-selected best/worst predictions}. Scores are mIoU (\%); random search reports mean $\pm$ standard deviation over three seeds.}
 \label{tab:configuration-cost}
 \begin{tabular*}{\columnwidth}{@{\extracolsep{\fill}}l*{4}{c}@{}}
 \toprule
 \textbf{Setting (cost)} & \textbf{LoveDA} & \textbf{OEM} & \textbf{UDD5} & \textbf{Vaihingen}\\
 \midrule
 AdaSeg ($1\times$) & 49.5 & 49.3 & 75.5 & 69.5\\
 Operator ALL ($1\times$) & 43.9 & 33.4 & 66.8 & 21.0\\
 \textcolor{gray}{R@2$^{+}$ ($2\times$, GT)} & \textcolor{gray}{$45.7\mathord{\pm}0.1$} & \textcolor{gray}{$35.8\mathord{\pm}0.3$} & \textcolor{gray}{$67.4\mathord{\pm}0.1$} & \textcolor{gray}{$29.0\mathord{\pm}0.5$}\\
 \textcolor{gray}{R@2$^{-}$ ($2\times$, GT)} & \textcolor{gray}{$42.3\mathord{\pm}0.1$} & \textcolor{gray}{$32.1\mathord{\pm}0.3$} & \textcolor{gray}{$65.9\mathord{\pm}0.8$} & \textcolor{gray}{$21.5\mathord{\pm}0.4$}\\
 \textcolor{gray}{R@10$^{+}$ ($10\times$, GT)} & \textcolor{gray}{$49.2\mathord{\pm}0.2$} & \textcolor{gray}{$41.4\mathord{\pm}0.2$} & \textcolor{gray}{$69.5\mathord{\pm}0.1$} & \textcolor{gray}{$42.8\mathord{\pm}1.3$}\\
 \textcolor{gray}{R@10$^{-}$ ($10\times$, GT)} & \textcolor{gray}{$39.0\mathord{\pm}0.1$} & \textcolor{gray}{$29.1\mathord{\pm}0.4$} & \textcolor{gray}{$65.0\mathord{\pm}0.9$} & \textcolor{gray}{$17.7\mathord{\pm}0.2$}\\
 \bottomrule
 \end{tabular*}
 \par\vspace{-5pt}
\end{table}

\begin{samepage}
\ablationheading{Configuration Selection and Execution Cost.} We compare AdaSeg's single policy call and selected execution with Operator ALL and random search (R@$K$) (\cref{tab:configuration-cost}).
Operator ALL enables all candidate descriptions and refinement operators; R@$K$ samples $K$ configurations from the starting configuration and candidate options.
Test GT selects the best and worst random prediction per image, providing an oracle unavailable during standard inference.
Execution counts measure configuration trials rather than wall-clock time, since runtimes can differ.
AdaSeg exceeds GT-selected R@10$^{+}$ by 0.3, 7.9, 6.0, and 26.7 points on LoveDA, OEM, UDD5, and Vaihingen, respectively.
Operator ALL falls below SegEarth-OV3 everywhere, showing that indiscriminately enabling operators can impair segmentation.
Increasing random trials from two to ten widens the best--worst gap on Vaihingen from 7.5 to 25.1 points.
More trials expose both beneficial and harmful configurations, while oracle selection remains limited to the sampled candidates.
AdaSeg conditions configuration generation on image evidence and the starting prediction.
Its advantage over GT-selected random search highlights the value of generating useful operator combinations before execution.\par
\end{samepage}

\subsection{Qualitative Results}
\label{sec:qualitative-results}
We compare predictions for the same validation images and target classes (\cref{fig:main-comparison}).
On LoveDA, AdaSeg reduces fragmented predictions within the large land-cover regions and better preserves their spatial continuity.
The Potsdam example shows more coherent building and surrounding land-cover predictions, with fewer isolated patches inside otherwise consistent regions.
On Vaihingen, AdaSeg corrects the class assignment of the large central roof, bringing its prediction closer to the ground truth.
The improvement involves correcting semantic labels as well as recovering object outlines.
All three examples use the same frozen segmenter with image-adaptive operator configurations.

\par
\makeatletter
\if@firstcolumn\newpage\fi
\makeatother
\section{Conclusion}
\label{sec:conclusion}

In this paper, we present \method, an image-adaptive framework for open-vocabulary remote sensing semantic segmentation.
Its vision-language policy learns from FLAIR-Operator to select a complete operator configuration, including text descriptions, fusion settings, and refinement operators, for a frozen segmenter in one call.
Execution-based optimization refines the source-supervised policy using rewards for segmentation quality, description coverage, and operator-response agreement.
Across eight benchmarks, it achieves 59.9\% average mIoU, 3.8 percentage points above the strongest frozen-segmenter baseline.
These results support adapting operator combinations to individual images as a complement to improving the segmentation model itself.
Future work will broaden source supervision and operator coverage toward reliable configurations across scenes and a general-purpose active perception framework.

\clearpage
\twocolumn
\raggedbottom
\section*{LLM Usage Statement}
We used large language models (LLMs) to assist with editing and debugging author-produced experimental code, including data processing, training, and evaluation scripts. All LLM-generated or modified code was reviewed, tested, and executed by the authors. LLMs also assisted with polishing the manuscript for grammar, clarity, and readability. All research ideas, hypotheses, method designs, experimental protocols, and scientific conclusions were developed by the authors. All reported numerical results originated from experiments rather than LLM generation. The authors take full responsibility for all content of this paper, including its claims, equations, citations, code, and results.

{
    \small
    \bibliographystyle{ieeenat_fullname}
    \bibliography{main}
}
\clearpage
\appendix
\setcounter{section}{0}
\setcounter{figure}{0}
\setcounter{table}{0}
\setcounter{equation}{0}
\renewcommand{\thesection}{\arabic{section}}
\renewcommand{\thefigure}{S\arabic{figure}}
\renewcommand{\thetable}{S\arabic{table}}
\renewcommand{\theHsection}{supp.\arabic{section}}
\renewcommand{\theHsubsection}{supp.\arabic{section}.\arabic{subsection}}
\renewcommand{\theHfigure}{supp.\arabic{figure}}
\renewcommand{\theHtable}{supp.\arabic{table}}
\renewcommand{\theequation}{S\arabic{equation}}
\renewcommand{\theHequation}{supp.\arabic{equation}}
\makeatletter
\def\@extra@b@citeb{-supp}
\renewcommand\NAT@wrout[5]{%
  \if@filesw
    {\let\protect\noexpand\let~\relax
     \immediate\write\@auxout{\string\bibcite{#5\@extra@b@citeb}{{#1}{#2}{{#3}}{{#4}}}}}\fi
  \ignorespaces}
\makeatother
\renewcommand{\dblfloatpagefraction}{0.5}
\raggedbottom
\maketitlesupplementary

\noindent
This supplementary material is organized as follows:
\begin{itemize}
 \setlength{\itemsep}{2pt}
 \setlength{\parskip}{0pt}
 \setlength{\parsep}{0pt}
 \item In \cref{sec:supp-datasets}, we describe the source and evaluation datasets, their label spaces, and the evaluation conventions.
 \item In \cref{sec:supp-flair-operator}, we explain how we construct FLAIR-Operator, covering candidate descriptions, operator probes, and source-supported target sampling.
 \item In \cref{sec:supp-implementation}, we detail the policy inputs, frozen segmenter execution, and settings for supervised fine-tuning and policy optimization.
 \item In \cref{sec:supp-analysis}, we compare policy models, examine operator strengths and interactions, and analyze configuration transfer and hyperparameter sensitivity.
 \item In \cref{sec:supp-qualitative}, we compare AdaSeg and SegEarth-OV3 on land-cover, object-rich aerial, and drone benchmarks.
\end{itemize}

\section{Dataset Details}
\label{sec:supp-datasets}

FLAIR provides source supervision for constructing FLAIR-Operator and training the policy.
We evaluate on eight remote sensing datasets covering land-cover mapping, urban aerial imagery, and drone scenes.
The descriptions below follow the benchmark scope of SegEarth-OV~\citesupp{DBLP:conf/cvpr/LiLCBZM025} and summarize the imagery and label spaces.

\paragraph{Source Data.}
FLAIR~\citesupp{DBLP:conf/nips/GarioudGLWVPGW23} contains aerial land-cover imagery acquired across France.
The dataset provides pixel-level annotations for $512\times512$ patches at a ground sampling distance of 0.2\,m.
We use RGB inputs and 13 classes covering built surfaces, roads, bare land, water, vegetation, and agricultural land.
We use these annotations to probe operators, measure segmentation quality, and provide execution feedback during policy training.
All configurations for an image belong to the same split. We exclude the image itself when estimating support from its neighbors.

\paragraph{Land-Cover Benchmarks.}
LoveDA~\citesupp{DBLP:conf/nips/WangZMLZ21} contains $1024\times1024$ Google Earth images at 0.3\,m resolution from urban and rural areas in three Chinese cities.
Its label space includes building, road, water, barren land, forest, agriculture, and background.
Urban and rural scenes share all seven labels and vary in object density and land-cover composition.
OpenEarthMap (OEM)~\citesupp{DBLP:conf/wacv/XiaYAB23} combines aerial and satellite imagery at spatial resolutions of 0.25--0.5\,m from 97 regions across 44 countries.
Its annotations distinguish bare land, rangeland, developed space, road, tree, water, agricultural land, and building.

\paragraph{Object-Rich Aerial Benchmarks.}
iSAID~\citesupp{DBLP:conf/cvpr/ZamirAGKSK00XB19} provides dense annotations for 15 object categories in DOTA-v1.0 imagery.
These include ships, aircraft, large and small vehicles, storage tanks, bridges, and sports facilities.
The semantic segmentation vocabulary comprises 16 labels, including background.
Objects are densely arranged and vary substantially in size and orientation.
Potsdam and Vaihingen are urban benchmarks from the ISPRS 2D Semantic Labeling Contest.
Their original ground sampling distances are 5\,cm and 9\,cm, respectively.
Both distinguish impervious surfaces, buildings, low vegetation, trees, cars, and clutter/background.
We evaluate the six classes separately for each benchmark using its class mappings and valid-pixel masks.

\paragraph{Drone Benchmarks.}
UAVid~\citesupp{LYU2020108} consists of high-resolution urban video sequences captured from oblique UAV viewpoints.
We process each frame independently of the other frames in its sequence.
We merge moving and static cars into one class and evaluate seven labels (background, building, road, car, tree, low vegetation, and human).
UDD5~\citesupp{DBLP:conf/prcv/ChenWLCW18} contains urban imagery captured by a DJI Phantom~4 at flight altitudes of approximately 60--100\,m.
Its five labels are vegetation, building, road, vehicle, and other/background.
VDD~\citesupp{DBLP:journals/jvcir/CaiJHGWY25} contains 400 RGB images of $4000\times3000$ pixels captured by a DJI Mavic Air~2 at approximately 50--120\,m altitude.
It covers urban, industrial, rural, and natural scenes with varied viewpoints and lighting.
Its seven labels are facade, roof, road, vegetation, vehicle, water, and other/background.

\paragraph{Evaluation Conventions.}
We compare AdaSeg with its starting operator configuration using identical evaluation images or windows, class definitions, and valid pixels.
To compute dataset-level mIoU, we pool class intersections and unions across the evaluation set.
The overall average weights all eight benchmarks equally. We compute per-image gains separately from the aggregate scores.
We use target annotations only to compute evaluation scores.

\vspace{-4pt}
\section{FLAIR-Operator}
\label{sec:supp-flair-operator}
\begin{figure}[!ht]
 \centering
 \includegraphics[width=\columnwidth]{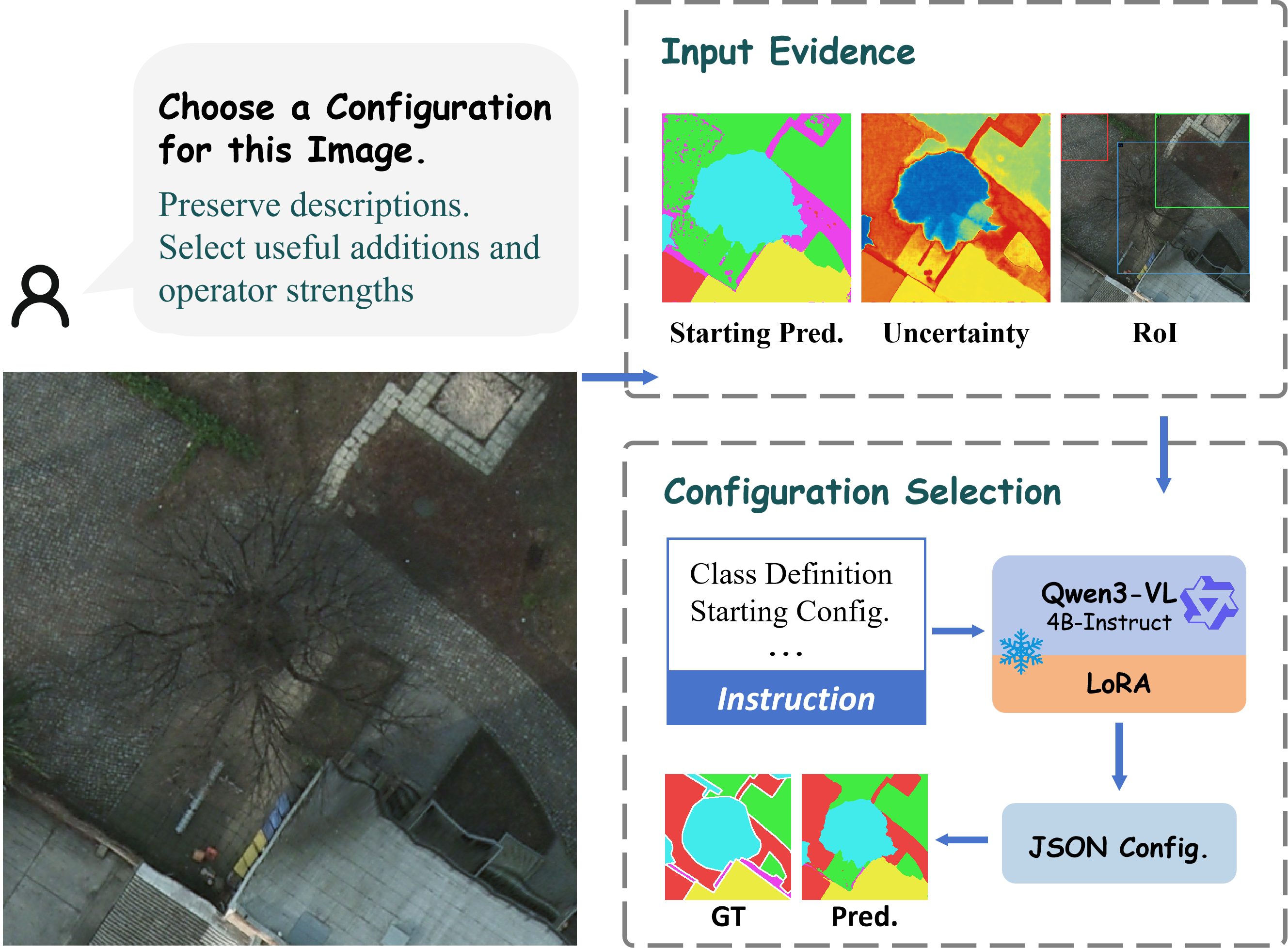}
 \par\vspace{0.000081388784\columnwidth}
 \caption{\textbf{Policy Input and Output.} Visual evidence comprises RGB, the starting prediction, continuous uncertainty, and crop regions. The policy generates a JSON configuration for execution by frozen SAM~3. Ground truth is shown for comparison.}
 \label{fig:prompt}
 \vspace{2pt}
\end{figure}

FLAIR-Operator pairs evidence with a target operator configuration.
The target specifies class descriptions for the instance, semantic, and presence heads, along with fusion settings, a background rule, and refinement operator strengths.
Sampled targets retain the starting descriptions and fusion settings and append selected candidate descriptions to the corresponding classes and heads.
At inference, the policy can add descriptions outside the candidate set while preserving the starting descriptions and target classes.
\Cref{fig:prompt} illustrates the JSON configuration the policy outputs for execution by the frozen segmenter.

\FloatBarrier

Operator ALL retains the starting descriptions, adds all candidates to their assigned classes and heads, and enables all refinement operators.
Text expansion can improve or reduce mIoU depending on the image.

Each probe varies one description or refinement setting and measures the per-image mIoU change from its paired reference (pp).
We retest choices with nonpositive standalone gains in sampled joint configurations. The reference omits the tested description or disables the tested operator.

We map the largest measured gain for each choice to a bounded response and assign zero to nonpositive gains.
Support is a distance-weighted average of responses from similar FLAIR images, excluding the current image.
The shared exploration coefficient $\eta$ mixes the source-supported distribution $P_{\mathrm{src}}$ with the uniform distribution $U$ for each decision.
\begin{equation}
 P=(1-\eta)P_{\mathrm{src}}+\eta U.
 \label{eq:supp-operator-sampling}
\end{equation}
We sample descriptions independently with probabilities given by their support.
The maximum support across strengths sets each operator's activation probability.
For an enabled operator, we sample strength in proportion to support, using a uniform distribution when all strengths have zero support.
Supervised training and replay draw a fresh target for each presentation and retain the starting descriptions and fusion settings.

\section{Implementation Details}
\label{sec:supp-implementation}

\subsection{Model and Execution Settings}
AdaSeg uses Qwen3-VL-4B-Instruct as its policy backbone and frozen SAM~3 as its segmenter (\Cref{tab:supp-settings}).
For each image or evaluation window, SAM~3 executes the starting configuration to produce the starting prediction.
The policy receives RGB, the starting prediction, a continuous uncertainty map, and RGB annotated with three crop regions selected under fixed reference descriptions.
The image-side limit applies to each evidence image. The output limit applies to the JSON configuration generated in one policy call.
SAM~3 then executes the selected configuration to produce the final segmentation.

\addvspace{8pt}
\noindent
\begin{minipage}{\columnwidth}
 \centering
 \captionsetup{hypcap=false}
 \captionof{table}{\textbf{Model, input, and execution settings.} \emph{Source neighbors} is the number of other FLAIR images used to estimate support. Crop scales are relative to the input image size.}
 \label{tab:supp-settings}
 \fontsize{8}{9.5}\selectfont
 \setlength{\tabcolsep}{3pt}
 \renewcommand{\arraystretch}{1.08}
 \begin{tabular*}{0.90\linewidth}{@{\extracolsep{\fill}\hspace{10pt}}ll@{\hspace{10pt}}}
 \toprule
 Setting & Value\\
 \midrule
 Policy & Qwen3-VL-4B-Instruct\\
 Frozen segmenter & SAM~3\\
 LoRA rank / alpha & 16 / 32\\
 Readout dim. & 128\\
 Visual evidence & \begin{tabular}[t]{@{}l@{}}
   RGB\\
   Starting prediction\\
   Uncertainty map\\
   Crop-annotated RGB
 \end{tabular}\\
 Max. image side & 672 px\\
 Max. output length & 2,048 tokens\\
 Policy calls & 1 per image or window\\
 Source pool & FLAIR\\
 Source neighbors & 32\\
 Probe strengths & 0.1, 0.25, 0.5\\
 Crop scales & 0.25, 0.5, 0.7\\
 \bottomrule
 \end{tabular*}
\end{minipage}

\subsection{Training Settings}
\Cref{tab:supp-training-settings} summarizes the settings for supervised fine-tuning (SFT) and group relative policy optimization (GRPO).
SFT trains the policy on the source-supported configuration targets described in Sec.~\ref{sec:supp-flair-operator}.
GRPO starts from the supervised policy and samples a group of configurations for each selected FLAIR image.
We execute each valid configuration with the same frozen segmenter and measure its mIoU gain over the starting configuration.
The reward weight balances the measured gain and agreement with the source-supported operator distribution.
The gate suppresses positive credit when the mIoU drop exceeds the regression tolerance.
KL regularization constrains deviation from the supervised policy. Replay draws fresh source-supported training targets.

\addvspace{8pt}
\noindent
\begin{minipage}{\columnwidth}
 \centering
 \captionsetup{hypcap=false}
 \captionof{table}{\textbf{Training settings.} The rollout count is per source image. The mIoU tolerance $\tau$ is measured in percentage points.}
 \label{tab:supp-training-settings}
 \fontsize{8}{9.5}\selectfont
 \setlength{\tabcolsep}{3pt}
 \renewcommand{\arraystretch}{1.08}
 \begin{tabular*}{0.90\linewidth}{@{\extracolsep{\fill}\hspace{8pt}}p{0.34\linewidth}*{2}{>{\centering\arraybackslash}p{0.20\linewidth}}@{\hspace{8pt}}}
 \toprule
 Setting & SFT & GRPO\\
 \midrule
 Optimizer & AdamW & AdamW\\
 Learning rate & $5\times10^{-5}$ & $1\times10^{-5}$\\
 Weight decay & 0.01 & 0.01\\
 Rollouts per image & -- & 8\\
 Reward weight $\alpha$ & -- & 0.2\\
 mIoU tolerance $\tau$ (pp) & -- & 0\\
 KL weight & -- & 0.04\\
 Replay weight & -- & 0.5\\
 Hardware & \multicolumn{2}{c@{\hspace{8pt}}}{$4\times$A100}\\
 \bottomrule
 \end{tabular*}
\end{minipage}

\begin{table*}[!t]
 \centering
 \fontsize{8}{9.5}\selectfont
 \setlength{\tabcolsep}{2pt}
 \renewcommand{\arraystretch}{1.15}
 \sisetup{
   mode=text,
   reset-text-series=false,
   table-number-alignment=center,
   table-column-width={\dimexpr(\textwidth-0.20\textwidth-24\tabcolsep)/12\relax}
 }
 \caption{\textbf{Comparison of policy models.} Exec. is the executable-output rate. Rep. is the share using the most frequent configuration. \colorbox[HTML]{FFF0CC}{\strut Starting config. copied} marks reuse of the starting configuration on every image. $^{*}$ marks mIoU after manual JSON correction. The \textcolor{red}{best} and \textcolor{blue}{second-best} mIoU are highlighted. Metrics are percentages, and mIoU covers all samples.}
 \label{tab:model-results}
 \begin{tabular}{@{}p{0.20\textwidth}*{4}{
   S[table-format={2.1{\textsuperscript{*}}},table-align-text-after=false]
   S[table-format=3.1]
   S[table-format=3.1]}@{}}
 \toprule
 \multirow{2}{*}{\textbf{Model}} & \multicolumn{3}{c}{\textbf{LoveDA}} & \multicolumn{3}{c}{\textbf{OEM}} & \multicolumn{3}{c}{\textbf{UDD5}} & \multicolumn{3}{c}{\textbf{Vaihingen}}\\
 \cmidrule(lr){2-4}\cmidrule(lr){5-7}\cmidrule(lr){8-10}\cmidrule(lr){11-13}
 & {mIoU$\uparrow$} & {Exec.$\uparrow$} & {Rep.$\downarrow$} & {mIoU$\uparrow$} & {Exec.$\uparrow$} & {Rep.$\downarrow$} & {mIoU$\uparrow$} & {Exec.$\uparrow$} & {Rep.$\downarrow$} & {mIoU$\uparrow$} & {Exec.$\uparrow$} & {Rep.$\downarrow$}\\
 \midrule
 Gemma3-4B~\citesupp{DBLP:journals/corr/abs-2503-19786} & 11.7{\textsuperscript{*}} & 29.2 & 50.5 & 38.4{\textsuperscript{*}} & 97.7 & 67.3 & \cellcolor[HTML]{FFF0CC} \color{blue} 71.7{\textsuperscript{*}} & \cellcolor[HTML]{FFF0CC} 100.0 & \cellcolor[HTML]{FFF0CC} 100.0 & 11.7{\textsuperscript{*}} & 15.3 & 64.1\\
 InternVL3.5-4B~\citesupp{DBLP:journals/corr/abs-2508-18265} & 39.1 & 81.4 & 6.1 & 14.9 & 34.1 & 22.6 & 31.7 & 40.0 & 7.5 & 4.8 & 8.3 & 11.8\\
 Qwen2.5-VL-3B~\citesupp{bai2025qwen25vltechnicalreport} & \color{blue} 48.3{\textsuperscript{*}} & 100.0 & 83.7 & 14.4{\textsuperscript{*}} & 45.8 & 24.6 & 3.2{\textsuperscript{*}} & 5.0 & 5.0 & 0.8{\textsuperscript{*}} & 1.0 & 11.6\\
 Qwen3.5-4B~\citesupp{qwen35blog} & 3.5 & 4.4 & 0.8 & 25.7{\textsuperscript{*}} & 65.6 & 18.0 & 55.1 & 100.0 & 72.5 & 42.4{\textsuperscript{*}} & 62.3 & 1.3\\
 Qwen3-VL-4B~\citesupp{bai2025qwen3vltechnicalreport} & \cellcolor[HTML]{FFF0CC} 47.4 & \cellcolor[HTML]{FFF0CC} 100.0 & \cellcolor[HTML]{FFF0CC} 100.0 & \cellcolor[HTML]{FFF0CC} \color{blue} 42.9 & \cellcolor[HTML]{FFF0CC} 100.0 & \cellcolor[HTML]{FFF0CC} 100.0 & \cellcolor[HTML]{FFF0CC} \color{blue} 71.7 & \cellcolor[HTML]{FFF0CC} 100.0 & \cellcolor[HTML]{FFF0CC} 100.0 & \cellcolor[HTML]{FFF0CC} \color{blue} 60.8 & \cellcolor[HTML]{FFF0CC} 100.0 & \cellcolor[HTML]{FFF0CC} 100.0\\
 \textbf{AdaSeg} & \color{red} 49.5 & 100.0 & 3.1 & \color{red} 49.3 & 100.0 & 1.1 & \color{red} 75.5 & 100.0 & 5.0 & \color{red} 69.5 & 100.0 & 6.8\\
 \bottomrule
 \end{tabular}
\end{table*}

\section{Additional Analysis}
\label{sec:supp-analysis}

\subsection{Policy Model Comparison}
\Cref{tab:model-results} compares policies with the same frozen executor per dataset.
AdaSeg led in mIoU on all four benchmarks with 100.0\% executability and 1.1--6.8\% repetition.
The original Qwen3-VL-4B always returned the starting configuration, yielding executable outputs at the starting mIoU.
Repetition thus helps identify policies that reuse a fixed configuration despite producing executable outputs.
With the executor fixed, AdaSeg's gains reflect image-adaptive configuration selection.

\begin{figure}[!tb]
 \centering
 \includegraphics[width=\columnwidth]{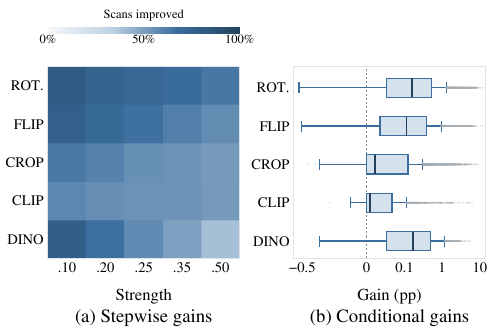}
 \caption{\textbf{Operator Strengths and Gains.} (a) Improvement rates relative to the preceding strength. (b) Conditional gains within joint configurations (pp). The dashed line marks zero gain.}
 \label{fig:strength-effects}
 \vspace{3pt}
\end{figure}

\begin{figure*}[!b]
 \centering
 \includegraphics[width=0.90\textwidth]{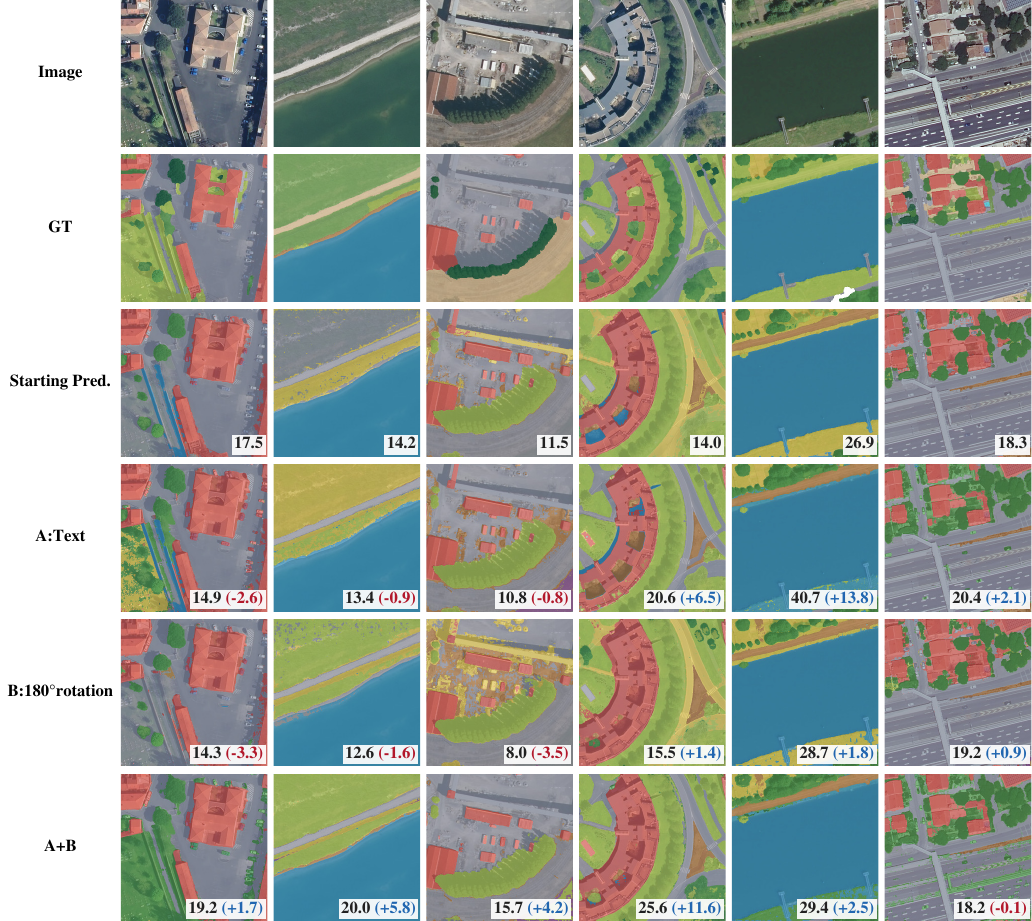}
 \caption{\textbf{Operator Interactions.} A denotes text expansion and B denotes $180^\circ$ rotation. The scores report mIoU (\%) over all classes.}
 \label{fig:supp-operator-interactions}
\end{figure*}

\subsection{Operator Strengths}
\Cref{fig:strength-effects} compares strengths from $0.05$ to $0.50$. The additional levels $0.05$, $0.20$, and $0.35$ are used only in this analysis.
Increasing strength improves only a subset of images, and that share varies by operator. Larger strengths generally help fewer images.
The best tested strength can thus differ across images even for the same operator, motivating image-adaptive selection of each operator's strength.

\subsection{Operator Interactions}
Figure~2(b) in the main paper shows that joint gains can depart from the sum of individual gains.
\Cref{fig:supp-operator-interactions} illustrates this for text expansion and $180^\circ$ rotation.
On the first three FLAIR images, each operator reduced mIoU alone, yet their combination improved it by 1.7, 5.8, and 4.2\,pp.
In the last two, both operators helped alone, while text expansion alone outperformed the pair.
Single-operator tests therefore cannot establish the benefit of a combination.
Selecting operators jointly for each image can recover useful combinations and avoid pairings that weaken individual gains.

\begin{table}[!t]
 \centering
 \caption{\textbf{Transfer of sampled configurations.} For unmatched target classes, \emph{Class name} uses the class label and \emph{All descriptions} uses the full description set from Operator ALL. All values are mIoU (\%).}
 \label{tab:supp-source-transfer}
 \fontsize{8}{9.5}\selectfont
 \setlength{\tabcolsep}{3pt}
 \renewcommand{\arraystretch}{1.08}
 \begin{tabular*}{0.90\linewidth}{@{\extracolsep{\fill}\hspace{8pt}}lcccc@{\hspace{8pt}}}
 \toprule
 \textbf{Fallback Setting} & \textbf{LoveDA} & \textbf{OEM} & \textbf{Vaihingen} & \textbf{UDD5}\\
 \midrule
 Class name & 40.6 & 43.1 & 51.3 & 71.5\\
 All descriptions & 38.4 & 38.3 & 45.4 & 66.3\\
 \textbf{AdaSeg} & \textbf{49.5} & \textbf{49.3} & \textbf{69.5} & \textbf{75.5}\\
 \bottomrule
 \end{tabular*}
\end{table}

\subsection{Transfer of Source-Sampled Configurations}
We tested cross-domain transfer with configurations sampled from a subset of FLAIR-Operator.
On this subset, the sampled configurations raised mIoU from 21.2\% to 33.6\% over the starting configuration.
Both variants reused sampled descriptions for matched classes.
For unmatched classes, they used the class name or the full Operator ALL description set.
We compared both with AdaSeg on four target benchmarks (\Cref{tab:supp-source-transfer}).

\smallskip
The class-name fallback outperformed the full description set on all four benchmarks, and AdaSeg improved on it by 4.1--18.2\,pp.
Differences in imagery and annotation conventions can change which descriptions and refinement operators are effective.
These shifts make it difficult to derive a single configuration from the source that works across target datasets, favoring selection based on each target image.

\newpage
\noindent
\begin{minipage}{\columnwidth}
 \centering
 \captionsetup{hypcap=false}
 \captionof{table}{\textbf{Hyperparameter sensitivity on Vaihingen.}
 Defaults are $\alpha=0.2$, $\tau=0$, KL weight $0.04$, and $\lambda_{\mathrm{rep}}=0.5$.
 Each ablation changes one hyperparameter and keeps the others at their defaults.
 Change reports the Non-declined / Declined image shares by per-image mIoU versus SegEarth-OV3. Results are percentages.
 Positive credit requires an mIoU gain of at least $-\tau$ percentage points over the starting configuration.}
 \label{tab:supp-grpo-sensitivity}
 \fontsize{8}{9.5}\selectfont
 \setlength{\tabcolsep}{3pt}
 \renewcommand{\arraystretch}{1.08}
 \begin{tabular*}{0.90\linewidth}{@{\extracolsep{\fill}\hspace{8pt}}llcc@{\hspace{8pt}}}
 \toprule
 \textbf{Factor} & \textbf{Value} & \textbf{mIoU} & \textbf{Change}\\
 \midrule
 AdaSeg & Default & \textbf{69.5} & \textbf{87.7 / 12.3}\\
 \midrule
 Reward weight $\alpha$ & $0.0$ & 68.0 & 71.6 / 28.4\\
          & $0.5$ & 68.5 & 71.4 / 28.6\\
          & $1.0$ & 68.4 & 68.1 / 31.9\\
 \midrule
 mIoU tolerance $\tau$ & $-0.2$ & 68.5 & 73.4 / 26.6\\
        & $+0.2$ & 67.8 & 71.1 / 28.9\\
        & $+0.5$ & 68.6 & 70.6 / 29.4\\
 \midrule
 KL weight & $0$ & 68.2 & 69.8 / 30.2\\
 \midrule
 Replay weight $\lambda_{\mathrm{rep}}$ & $0$ & 68.1 & 67.8 / 32.2\\
 \bottomrule
 \end{tabular*}
\end{minipage}

\newpage
\subsection{Hyperparameter Sensitivity}
The default led on Vaihingen with 69.5\% mIoU and 12.3\% declines (\cref{tab:supp-grpo-sensitivity}).
Both $\alpha$ endpoints lowered mIoU, favoring a mix of execution gains and source-supported operator agreement.
Changing $\tau$ or removing KL or replay increased declines.
The stricter gate ($\tau=-0.2$) also lowered mIoU, favoring the zero threshold on both metrics.
The KL and replay ablations suggest that continued supervision helps limit per-image regressions.

\section{Additional Qualitative Results}
\label{sec:supp-qualitative}
\begin{figure*}[p]
  \centering
  \includegraphics[width=0.98\textwidth]{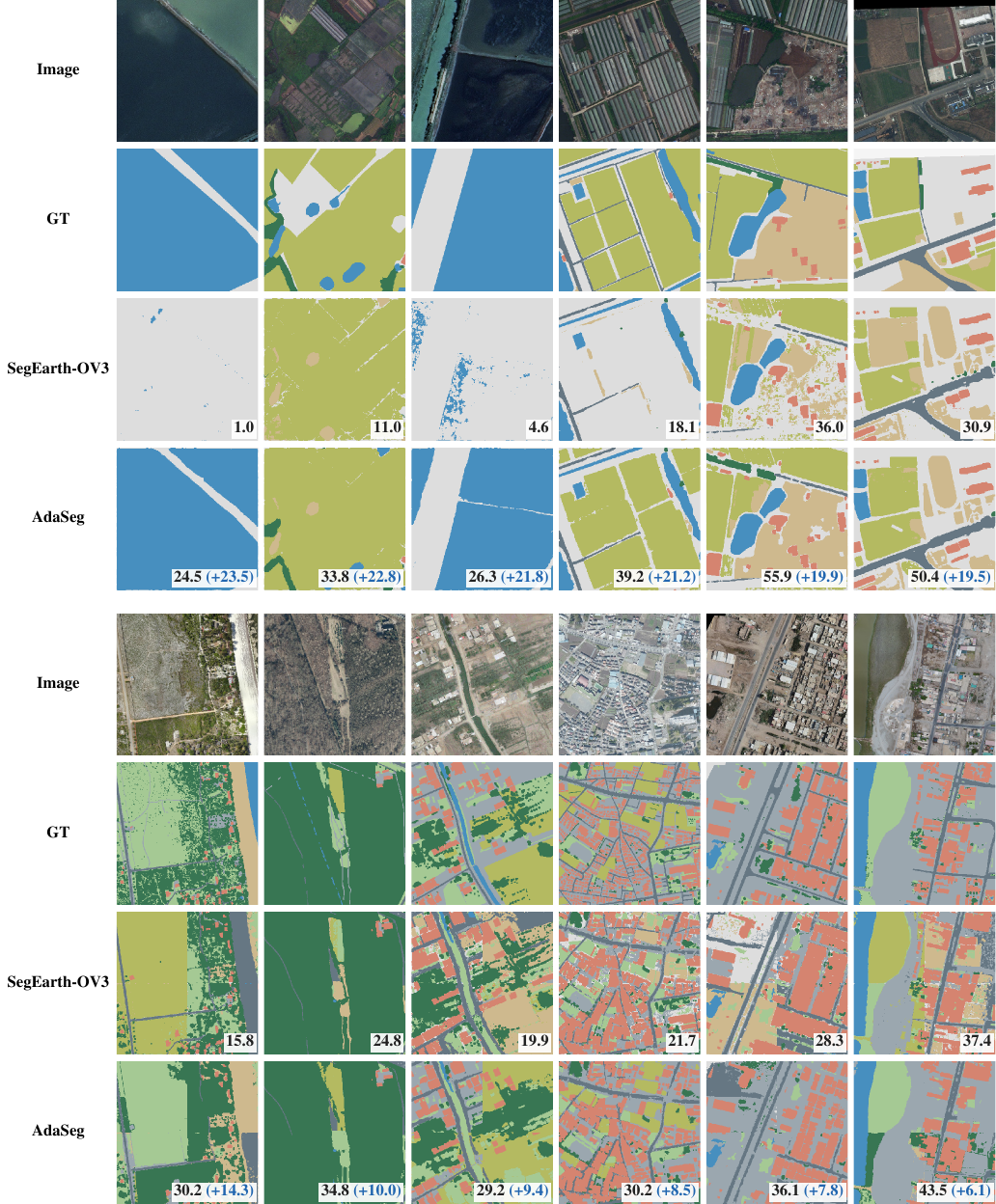}
  \caption{\textbf{Qualitative Results on Land-Cover Benchmarks.} Scores are mIoU (\%) over all classes.}
  \label{fig:qualitative-land-cover}
\end{figure*}

\begin{figure*}[p]
  \centering
  \includegraphics[width=0.98\textwidth]{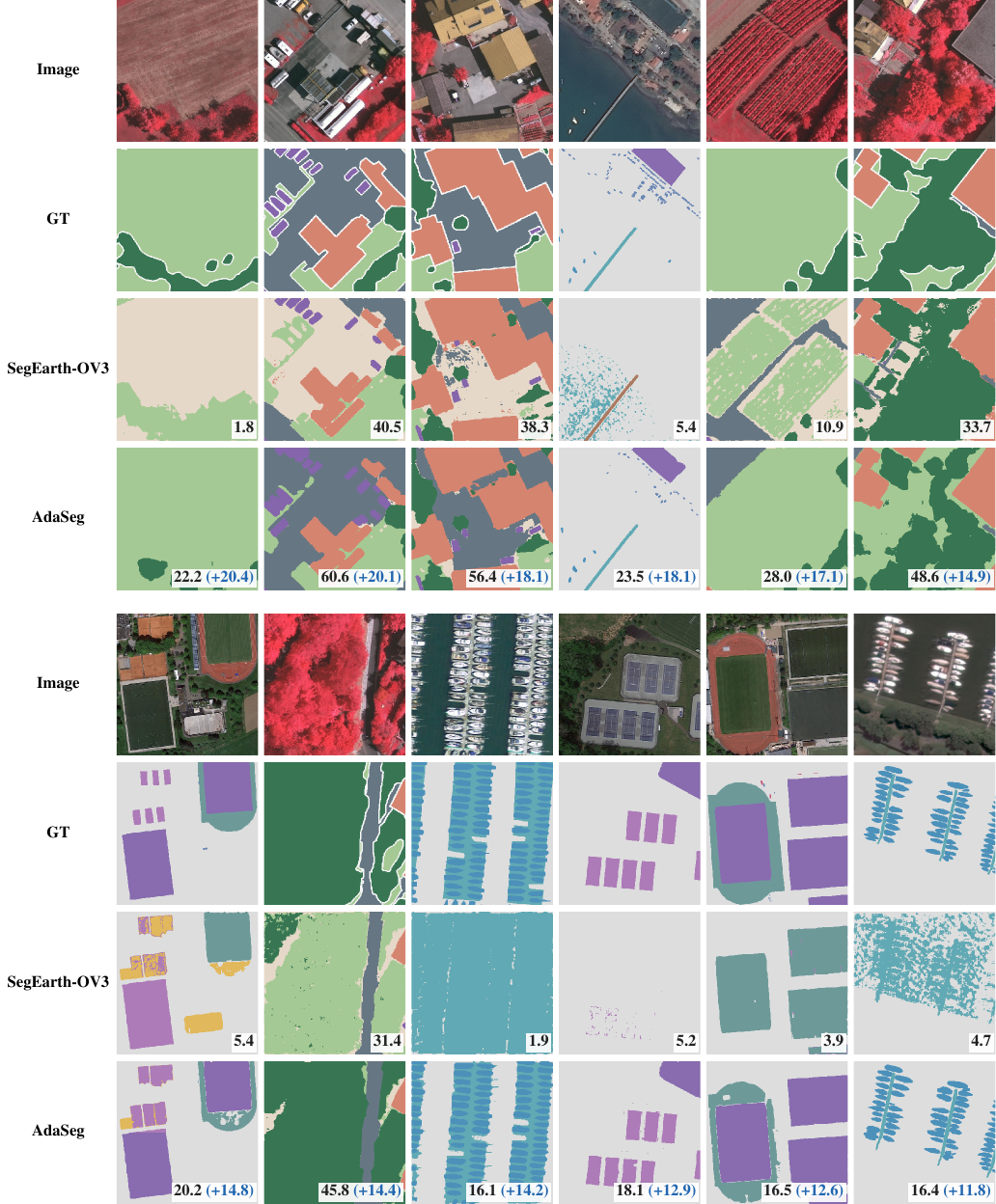}
  \caption{\textbf{Qualitative Results on Object-Rich Aerial Benchmarks.} Scores are mIoU (\%) over all classes.}
  \label{fig:qualitative-object-rich}
\end{figure*}

\begin{figure*}[p]
  \centering
  \includegraphics[width=0.98\textwidth]{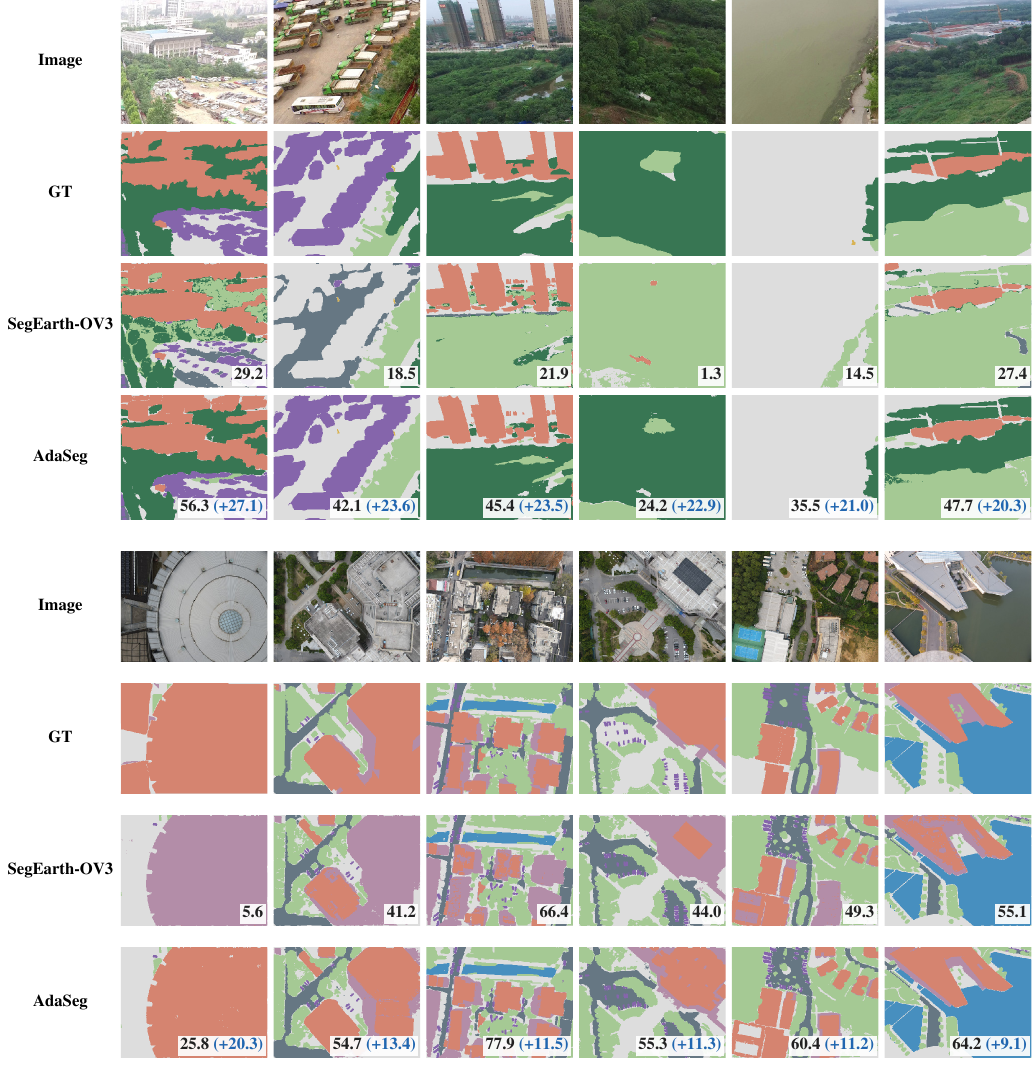}
  \caption{\textbf{Qualitative Results on Drone Benchmarks.} Scores are mIoU (\%) over all classes.}
  \label{fig:qualitative-drone}
\end{figure*}

\Cref{fig:qualitative-land-cover,fig:qualitative-object-rich,fig:qualitative-drone} show results on land-cover, object-rich aerial, and drone benchmarks.
Each column pairs a scene's RGB image and ground truth with predictions from SegEarth-OV3 and AdaSeg.
Scores report per-image mIoU over all classes.
In these examples, AdaSeg yields more coherent regions and sharper object boundaries.

\clearpage
\makeatletter
{
 \small
 \bibliographystylesupp{ieeenat_fullname}
 \if@filesw\immediate\write\@auxoutsupp{\string\bibdata{supp}}\fi
 \@input@{supp.bbl}
}
\makeatother

\end{document}